\documentclass[10pt,twocolumn,letterpaper]{article}

\usepackage{iccv}
\usepackage{times}
\usepackage{epsfig}
\usepackage{graphicx}
\usepackage{amsmath}
\usepackage{amssymb}

\usepackage{caption}
\usepackage{subcaption}

\usepackage[breaklinks=true,bookmarks=false]{hyperref}

\iccvfinalcopy % *** Uncomment this line for the final submission

\def\iccvPaperID{****} % *** Enter the ICCV Paper ID here

\ificcvfinal\fi

\begin{document}

%%%%%%%%% TITLE
\title{Guiding Query Position and Performing Similar Attention for Transformer-Based Detection Heads}

\author{Xiaohu Jiang\thanks{Work done during an internship at MEGVII Technology.}$^{~~1, 2}$\quad Ze Chen$^{2}$\quad Zhicheng Wang$^{2}$\quad Erjin Zhou$^{2}$\quad ChunYuan\thanks{Correspoding author.}$^{~~1}$\\
Tsinghua Shenzhen International Graduate School$^{1}$\\
MEGVII Technology$^{2}$\\
% Shenzhen, China\\
{\tt\small jiangxh21@mails.tsinghua.edu.cn\quad chenze@megvii.com}\\ 
{\tt\small wangzhicheng@megvii.com\quad zej@megvii.com\quad yuanc@sz.tsinghua.edu.cn}
}

\maketitle
\ificcvfinal\thispagestyle{empty}\fi

%%%%%%%%% ABSTRACT
\begin{abstract}

After DETR was proposed, this novel transformer-based detection paradigm which performs several cross-attentions between object queries and feature maps for predictions has subsequently derived a series of transformer-based detection heads. These models iterate object queries after each cross-attention. However, they don't renew the query position which indicates object queries' position information. Thus model needs extra learning to figure out the newest regions that query position should express and need more attention. To fix this issue, we propose the Guided Query Position (GQPos) method to embed the latest location information of object queries to query position iteratively.

Another problem of such transformer-based detection heads is the high complexity to perform attention on multi-scale feature maps, which hinders them from improving detection performance at all scales. Therefore we propose a novel fusion scheme named Similar Attention (SiA): besides the feature maps is fused, SiA also fuse the attention weights maps to accelerate the learning of high-resolution attention weight map by well-learned low-resolution attention weight map.

Our experiments show that the proposed GQPos improves the performance of a series of models, including DETR, SMCA, YoloS, and HoiTransformer and SiA consistently improve the performance of multi-scale transformer-based detection heads like DETR and HoiTransformer.

\end{abstract}

%%%%%%%%% BODY TEXT
\section{Introduction}

Recently, DETR~\cite{carion2020end} has proposed a novel detection paradigm based on transformer~\cite{vaswani2017attention} architecture. This kind of detection heads predict results by performing several multi-head attentions, named cross-attention (Eq.~\ref{cross_attn}), between object queries and feature maps and shows good performance on detection tasks. Therefore, many other works~\cite{gao2021fast, fang2021you} further studied this novel transformer-based detection paradigm and extended it to other vision tasks~\cite{zou2021end}. 

DETR and its derived models all update the object queries after each cross-attention, however, they don't renew the object queries' position encoding, named query position. They adopt an identical query position for every cross-attention layer, thus object queries' position information indicated by query position is not updated. Consider that in cross-attention, besides object queries, query position will also interact with the feature map to enhance the attention of the regions expressed by query position in the feature map. Thus not embedding object queries' latest location information to query position will take extra time for the model to learn the newest regions that query position should express and focus on.

To fix this issue, we propose GQPos method (sec~\ref{GQPos}): query position is updated after object queries iterate, which won't bring extra learning burden. To embed the newest location information of object queries to query position, predicted positions from the latest object queries are encoded, and are utilized as the query position of next cross-attention. Since query position is embedded with the newest location information, the model is free from extra learning. Cosine positional encoding is chosen to encode locations so that the learning procedure only brings little overhead. Visualization on DETR (Figure~\ref{fig:vis}) shows that after applying GQPos, the focused regions of DETR's decoder gradually transfer from the center to the boundary of the object.

Transformer-based detection heads also try to use multi-scale feature maps to improve detection on objects at all scales. And on classical CNN models, this can be achieved by fusing multi-scale features~\cite{lin2017feature}. However, it is difficult for the transformer to perform attention on high-resolution feature map, because the complexity of attention is quadratic in the number of pixels, so feature fusion alone has limited effects on transformer-based detection heads. Therefore, the Similar Attention (SiA) fusion scheme (sec~\ref{SiA:M}) is proposed. Considering that the distribution of attention maps at different scales should be similar, we propose that besides the feature maps can be fused, the multi-scale attention weight maps can also be fused. Specifically, the low-resolution attention weight map which is easier to learn is interpolated as the prior of high-resolution attention map. As a result, the learning of high-resolution attention map can be accelerated.

The effectiveness of GQPos and SiA is verified on two tasks, object detection and human-object-interaction. For object detection, when applied GQPos, DETR achieves $42.0$\% mAP at $50$ epochs under single-scale, which surpasses the original counterpart with a 10x training schedule. Furthermore, when combined with the proposed SiA method under multi-scale, the model achieves SOTA performance of $44.0$\% mAP with Resnet-$50$ backbone. Experiments on SMCA~\cite{gao2021fast} and YoloS~\cite{fang2021you} also show consistent improvements, which also verified the effectiveness of GQPos. For human-object-interaction, the single and multi-scale HoiTransformer's~\cite{zou2021end} performance can also get improved by applying GQPos and SiA respectively.

We hope our research will bring new thoughts about transformer-based detection heads' query position and attention weights fusion to the community. Our work provides the following three contributions:
\begin{itemize}
\item We analyze the mechanism of query position in transformer-based detection head and find out that the query position is not fully optimized to involve the latest location information. We proposed GQPos (sec~\ref{GQPos}) to embed the latest location information of object queries to query position iteratively.
\item Feature fusion has limited effects on transformer-based detection heads due to attention's high complexity, which hinders them from improving detection performance at all scales. Therefore, we propose SiA (sec~\ref{SiA:M}), which fuse multi-scale attention weight maps to accelerate the learning of high-resolution feature map by interpolating well-learned low-resolution feature map.
\item The proposed GQPos and SiA methods shows consistent improvement on object detection and human-object-interaction tasks. In particular, when applied GQPos, DETR surpasses the performance of the original version with a 10x training schedule. Furthermore, when combined with SiA, DETR achieves SOTA performance of $44.0$\% mAP under multi-scale.
\end{itemize}
%-------------------------------------------------------------------------

\begin{figure*}[h]
\centering
\includegraphics[width=1.\textwidth]{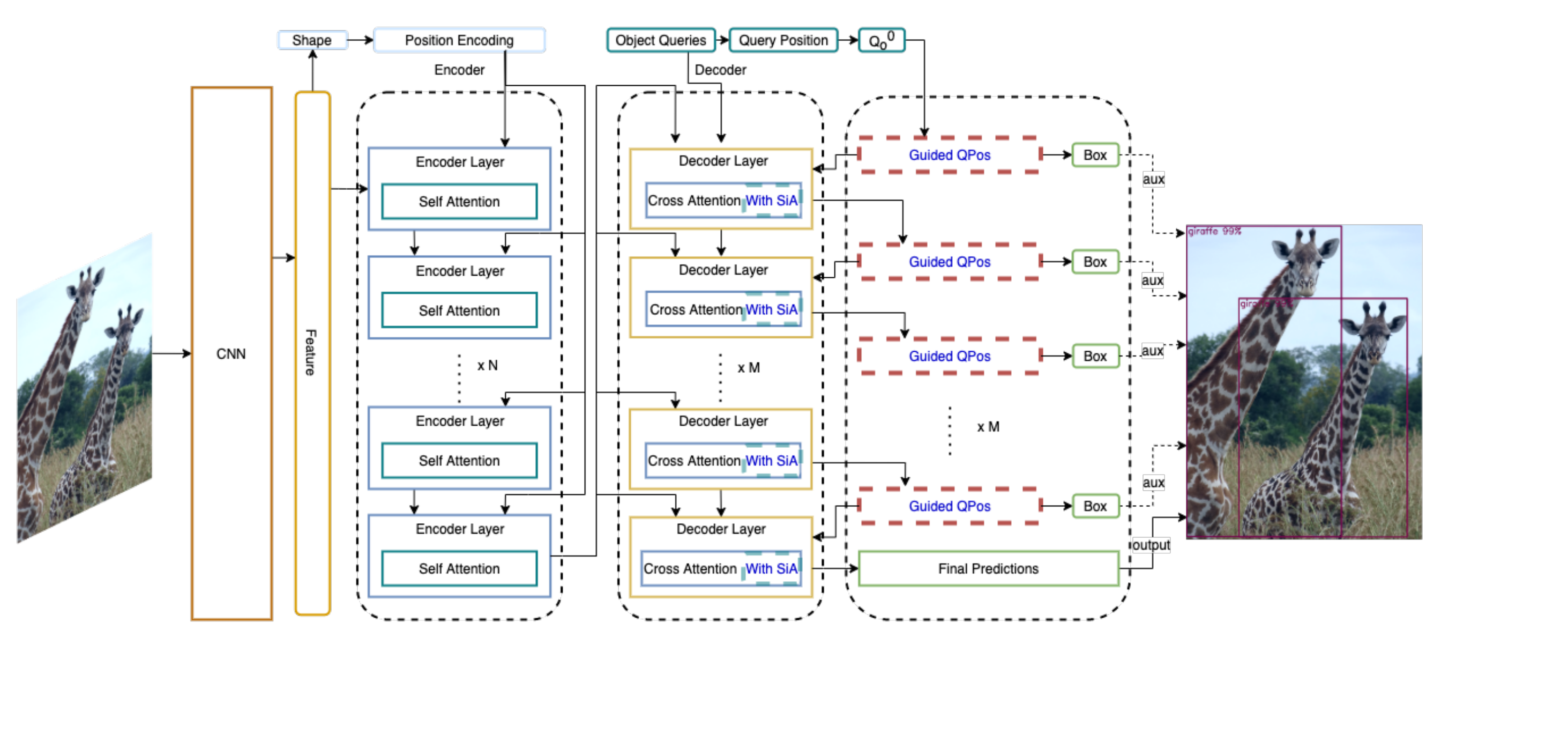} 
\caption{An overview architecture of DETR with GQPos and SiA. The features extracted by backbone are flattened and passed to the encoder for self-attention calculation, then the enhanced features are feed to the decoder to perform cross-attention with object queries which are embedded with query position. Decoder layers' outputs are used for predictions. We highlight the parts where our two methods are inserted. SiA is used in each decoder layer's cross-attention and GQPos utilizes the latest output object queries and passes the guided query position to the next layer. The details of the two methods are described in sec~\ref{GQPos} and sec~\ref{SiA:M}, respectively.}
\label{detr}
\end{figure*}

\section{Related Work}
\subsection{Object Detection}
In recent years, deep learning has been successfully applied to object detection~\cite{girshick2015region}. The object detection framework based on deep learning can be divided into several categories: two-stage and one-stage, or anchor-free and anchor-based. Recently, the research of end-to-end object detection has become a hot topic.

Common two-stage object detectors, such as RCNN~\cite{girshick2015region}, Fast RCNN~\cite{girshick2015fast} and Faster RCNN~\cite{ren2015faster}, extract candidate feature regions in the first stage and then object classification and location estimation are performed in the second stage. One stage detectors, such as Yolo~\cite{redmon2016you} and SSD~\cite{liu2016ssd}, perform object classification and location prediction directly on dense anchors. At present, the most popular two-stage and one-stage detectors contain complex hand-crafted components, such as NMS and anchors. The design of these components also has a great impact on the final detection results.

Recently, end-to-end object detectors have become more and more popular. They remove the complex post-processing such as NMS, and achieve one-to-one matching between the target and the candidate by the Hungarian algorithm~\cite{stewart2016end}. DETR~\cite{zhu2020deformable} and POTO~\cite{wang2020end} are some of successful examples. Moreover, DETR has successfully introduced transformer~\cite{vaswani2017attention} into the field of object detection, but it is limited by the slow convergence. 

To address the slow convergence issue of DETR, Deformable DETR~\cite{zhu2020deformable} propose that the high computational complexity of attention in the image field causes the slow convergence of DETR, especially the self-attention computational complexity of the encoder is the square of the number of image pixels, so it uses sparse sampling attention to accelerate the convergence. UP-DETR~\cite{dai2020up} uses self-supervised pre-training to improve DETR, while TSP-RCNN/FCOS~\cite{sun2020rethinking} thinks that the existing Hungarian matching algorithm and the cross-attention of decoder cause the slow convergence, and that decoder will destroy the performance of small objects. So it only adds the encoder to the origin FCOS~\cite{tian2019fcos} and Faster RCNN. Adaptive clustering transformer (ACT)~\cite{zheng2020end} proposes the idea of clustering to reduce the computational complexity of self-attention of the encoder. Recently, SMCA~\cite{gao2021fast} improves the sparse attention of Deformable DETR by generating a gaussian map centered on the reference point to enhance the attention near the reference point, compared with Deformable attention, it uses more global information. DETR's new detection paradigm is also used for tasks like human-object interaction, such as HoiTransformer~\cite{zou2021end}, which predicts people, objects, and their interaction in images. What's more, this paradigm has also been tried to be combined with ViT~\cite{dosovitskiy2020image}, such as YoloS~\cite{fang2021you}, YoloS adds 100 object queries to the origin patches of Vision Transformer to perform multi-head attention, and objects are predicted by these added object queries.

\subsection{Transformer In Vision}
Transformer structure is based on self-attention and cross-attention mechanisms to learn the relationship between the elements in the input sequence. Unlike RNN~\cite{mikolov2010recurrent} and other recurrent networks, which can only process the input sequence recursively and focus on the short-term context, the transformer can learn the long-term relationships of the input sequence and has a higher degree of parallelism.
One of the characteristics of transformer structure is that its scalability to high complexity models and large-scale datasets, and transformer does not need prior assumptions or knowledge about the problem compared to convolutional neural networks~\cite{krizhevsky2012imagenet} and recurrent networks. Therefore, the transformer can be well pre-trained on large-scale unlabeled datasets, and then it can be used in downstream tasks after fine-tuning to get the expected results.

The great success of transformer in the field of natural language processing (NLP) has aroused the interest of computer vision community. Transformer and its variants have been successfully used in image classification~\cite{dosovitskiy2020image, touvron2020training}, segmentation~\cite{ye2019cross}, image super resolution~\cite{yang2020learning}, object detection~\cite{carion2020end}, video understanding~\cite{sun2019videobert,girdhar2019video}, image generation~\cite{chen2020pre}, text image synthesis and visual question answering~\cite{tan2019lxmert,su2019vl}. DETR is a typical application of transformer in the field of object detection. In 
DETR, the feature information extracted by CNN is encoded by the encoder of the transformer. Then in the decoder, a series of object queries interact with these encoded features through cross-attention, to predict the locations and classes of objects in the image. 

\section{Method}
In this section, we introduce two novel methods, Guided query position (GQPos) and Similar Attention (SiA). An overall architecture of DETR with GQPos and SiA is shown in Figure~\ref{detr}. The extracted features are first enhanced by the encoder's self-attention, then the enhanced features are feed to decoder layers to perform cross-attention with object queries which are embedded with query position. Decoder layers' outputs are used for predictions. SiA is used in each cross-attention layer and GQPos iteratively guided query position.

\subsection{Guided Query Position}
\label{GQPos}
When cross-attention is performed between object queries and the input feature map, the information of feature map is embedded to object queries to make predictions. A typical cross-attention is as follows:
\begin{equation}
\begin{aligned}
    Q_{o}^{i} &= softmax(\frac{1}{\sqrt{d}}Q^{i-1}K^T)V \\
    &= softmax(\frac{1}{\sqrt{d}}(Q_{o}^{i-1}+Q_{pos})K^T)V \\
    &= softmax(\frac{1}{\sqrt{d}}(Q_{o}^{i-1}K^T + Q_{pos}K^T))V,
\label{cross_attn}
\end{aligned}
\end{equation}

\noindent where $Q_{o}^{i}$ is the output object queries of the $i$-th decoder layer, $K$ refers to the enhanced feature map output by the encoder, $d$ is the dimension of $K$. As the encoding of object queries, the query position $Q_{pos}$ indicates the location information of the predicted objects. The $Q_{o}K^{T}$ term enables the feature map's information to be embedded to object queries and the $Q_{pos}K^{T}$ term strengthens the attention of the regions expressed by query position in the feature map. Previous models ignore the update of $Q_{pos}K^{T}$ term thus model takes extra time to learn the latest regions query position should express.

Different from previous methods, the proposed GQPos update the query position during the iteration of cross-attention layers. The motivation of GQPos is that the update of query position should be guided to embed the latest position information of object queries. Thus we first calculate the locations of objects predicted by the output object queries of the former layer:
\begin{equation}
    pos^{i}(x,y,h,w) = MLP(Q_{o}^{i-1}),
\end{equation}

\begin{figure}[t]
\centering
\includegraphics[width=0.9\columnwidth]{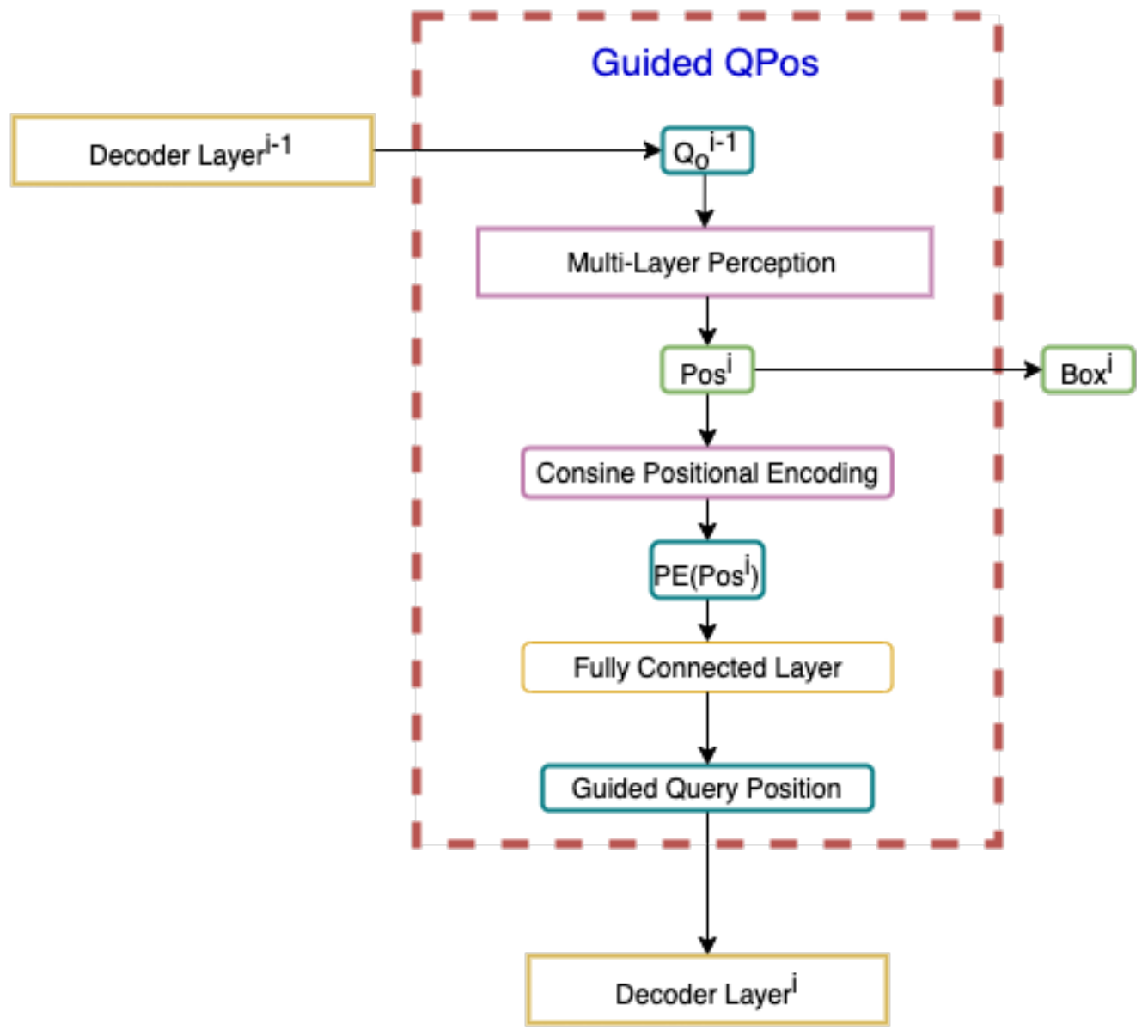} 
\caption{The structure of Guided query position method between $i-1$-th and $i$-th decoder layer.}
\label{gqpos}
\end{figure}

\noindent where $x,y,h,w$ are the center, height, width of the predicted object respectively. Then we use the cosine positional encoding to encode the $pos^{i}(x,y,h,w)$:
\begin{equation}
\begin{aligned}
 PE_{(pos^i,2j)} = sin(\frac{pos^i}{10000^{\frac{2j}{d_{model}}}}), \\ 
 PE_{(pos^i,2j+1)} = cos(\frac{pos^i}{10000^{\frac{2j}{d_{model}}}}),
\label{embed}
\end{aligned}
\end{equation}

\noindent where $j$ is the dimension. Each dimension of the positional encoding corresponds to a sinusoid. After that the location information of the latest object queries is embedded to $PE(pos^{i})$. Finally, we add a linear projection for $PE(pos^{i})$
% and the spatial position of the feature map
. The cross-attention for $i$-th layer is changed to:
\begin{equation}
\begin{aligned}
    Q_{o}^{i} = softmax(\frac{1}{\sqrt{d}}(Q_{o}^{i-1}+QPos^{i})K^{T})*V \\
    = softmax(\frac{1}{\sqrt{d}}(Q_{o}^{i-1}+FC(PE(pos^{i}))K^T)V,
\end{aligned}
\end{equation}

\noindent where $F$ is the enhanced feature map
% and $spos$ is the spatial position encoding for $F$
. The structure of GQPos is illustrated in Figure~\ref{gqpos}.

\begin{figure*}[h]
\centering
\includegraphics[width=.9\textwidth]{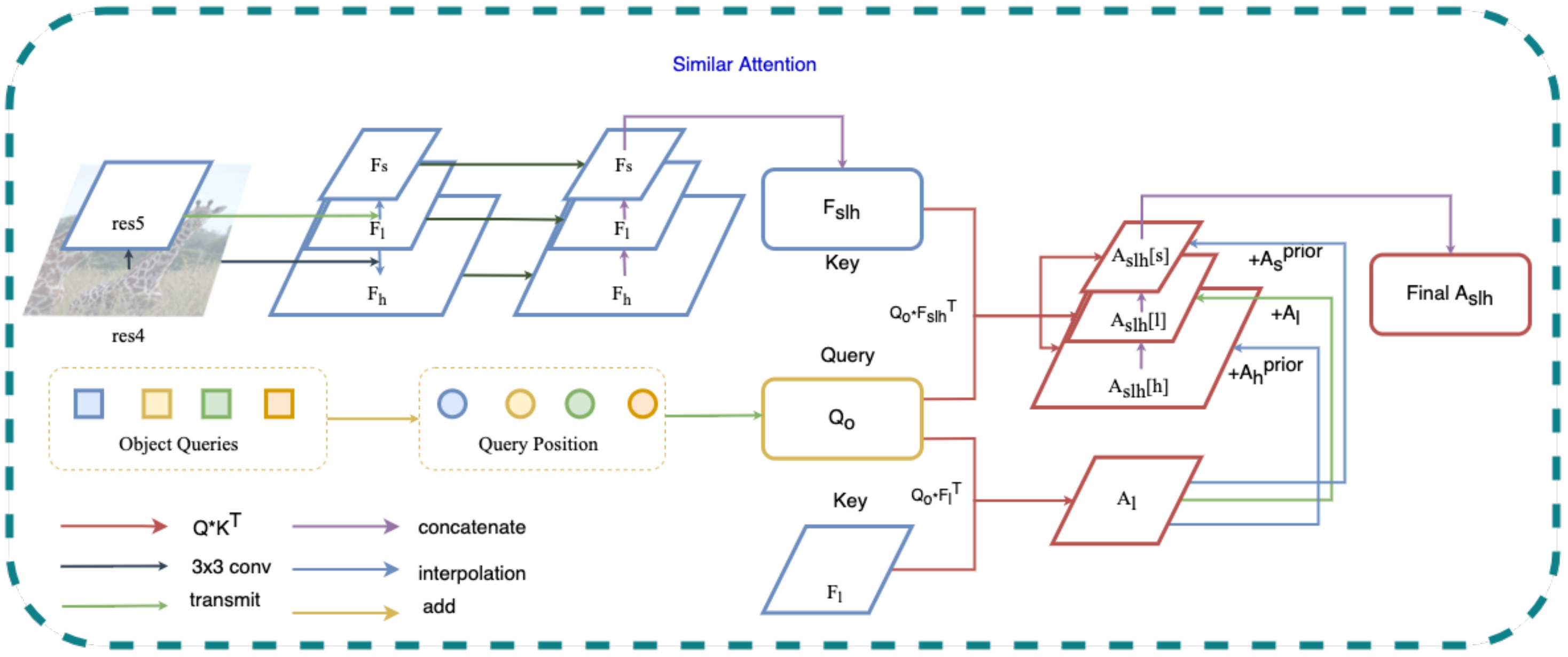} 
\caption{The structure of Similar Attention on DETR. It contains the part of feature map fusion and the part of attention weight fusion. We choose $F_{l}$ (res$5$) to interpolate the priors since $F_{l}$ is well learned in single-scale DETR.}
\label{SiA}
\end{figure*}

\subsection{Similar Attention}
\label{SiA:M}
As for transformer with multi-scale feature maps, the attention weight map is calculated as follows:
\begin{equation}
\begin{aligned}
    A_{l_h} = (Q_{o} +Q_{pos}) * F_{l_h}^{T},
\end{aligned}    
\end{equation}
\noindent where $F_{l_h}$ is the stitching of high resolution and low-resolution feature maps, and $A_{l_h}$ is the stitching attention weight map. The attention weight map $A_{l_h}$ contains the hard-to-learn high-resolution part and easy-to-learn low-resolution part.

Since the locations of the objects on different scale feature maps are similar, the distribution of attention weight maps at different scales should also be similar. Thus SiA interpolates the low-resolution attention weight map to get the prior of the high-resolution attention map. The prior is calculated as follows:
\begin{equation}
\begin{aligned}
    A_{l} = (Q_{o} +Q_{pos})* F_{l}^{T}\\
    A_{h}^{prior} = Bilinear(A_{l}),
\end{aligned}
\end{equation}

\noindent where $F_{l}$ is the low resolution feature map and $A_{h}^{prior}$ is the calculated prior. Finally SiA combines the prior and origin high resolution attention weight map:
\begin{equation}
\begin{aligned}
    A_{l_h}[h] = A_{l_h}[h] + \beta A_{h}^{prior}\\
    \beta = FC(Q_{o} + Q_{pos}),
\end{aligned}    
\end{equation}

\noindent where $A_{l_h}[h]$ represents the high-resoluation part in the stitching attention weight map and $\beta$ is a learnable coefficient. In our implementation, we add an extra small resolution feature map $F_{s}$ downsampled by $F_{l}$. The structure for Similar Attention is shown in Figure~\ref{SiA}. 

%------------------------------------------------------------------------
\section{Experiments}
\subsection{Datasets and metrics}
\noindent \textbf{COCO~\cite{lin2014microsoft}} We evaluate object detection performance on COCO $2017$ dataset. Specifically, we train on the COCO 2017 training data set and validate on the validation data set. The COCO $2017$ dataset contains $118K$ training images and $5K$ validation images respectively. We use mean average precision (mAP) to measure the performance of our method. \\
\noindent \textbf{HICO-DET~\cite{chao2015hico}} HOI detection is considered as true positive not only human and object is localized accurately but also the interaction has to be predicted correctly. We evaluate the performance of human-object-interaction task on HICO-DET dataset. There are $47,776$ images and more than $150K$ human-object pairs in HICO-DET with $600$ HOI categories over $117$ interactions and $80$ objects. HOI categories are split into $138$ Rare and $462$ Non-Rare based on the number of training instances. The training set includes $38,118$ images and the testing set includes $9,658$ images. The mAP is still used to examine the model performance.\\
\subsection{Implementation details}
All the experiments for single-scale models are trained on $8$ RTX $2080$ Ti GPUs and multi-scale models are trained on $8$ V$100$ GPUs.\\
\noindent \textbf{DETR and SMCA} For DETR and SMCA, We follow experimental settings of ~\cite{carion2020end}. The batch size per GPU is set to $1$. The learning rate of backbone is $1e-5$ and the learning rate of the transformer part is $1e-4$. We drop the learning rate to $1e-5$ in the $40$th epoch and experiments are based on the training results of $50$ epochs. The loss of classification is changed from cross-entropy loss to focal loss~\cite{lin2017focal}, the number of queries is replaced by $300$, and reference points is used in DETR's location regression to make fair comparisons with ~\cite{zhu2020deformable,gao2021fast}. Multi-head attention's std is decreased when initializing to stabilize the training process. The whole model is optimized by AdamW optimizer~\cite{loshchilov2018fixing} and pretrained with Resnet-$50$\cite{he2016deep}. Random cropping is used for data augmentation with larges width and height are setting as $1333$.\\
\noindent \textbf{YoloS} For YoloS, we choose YoloS-Ti model in our experiments. We follow the general settings of ~\cite{fang2021you}, but some modifications are made to the YoloS-Ti architecture, including adding more $6$ transformer layers since origin YoloS has no detection head part except a ViT backbone, and auxiliary losses are added to the extra $6$ layers to improve intermediate object queries' expressions. Model is pretrained with Deit-Ti~\cite{touvron2020training} and fine-tuned on COCO $2017$ with $50$ epochs.\\
\noindent \textbf{HoiTransformer} For HoiTransoformer, the experimental settings are basically the same with ~\cite{zou2021end}, and for the convenience of verification we train HoiTransformer using pretrained Resnet-$50$~\cite{he2016deep}. For single-scale HoiTransformer, the training procedure lasts for $30$ epochs, with the learning rate drop at the $20$-th epoch. For multi-scale HoiTransformer, we train $50$ epochs and drop the learning rate at $40$-th epochs.

\begin{table*}[t]
\begin{center}
\begin{tabular}{|l|c|c|c|c|c|c|c|c|c|}
\hline
Method & Epochs  & mAP &AP$_{50}$ &AP$_{75}$ &AP$_{s}$ & AP$_{M}$ & AP$_{L}$ & GFLOPs & Params(M)\\
\hline\hline
Faster RCNN-FPN-R50~\cite{ren2015faster}  & 36 & 40.2 & 61.0 & 43.8 & \textbf{24.2} & 43.5 & 52.0 & 180 & 42  \\
Deformable DETR-R50~\cite{zhu2020deformable}  & 50 & 39.4 & 59.6 & 42.2 & 20.6 & 43.0 & 55.5 &  78 & 34\\
SMCA-R50~\cite{gao2021fast} (2021) & 50 & 41.0 & * & * & 21.9 & 44.3 & 59.1 & 86 & 42 \\
DETR-GQPos-R50&  50 & \textbf{42.0} & \textbf{63.1} & \textbf{44.7} & 22.7 & \textbf{46.1} & \textbf{60.5} & 86 & 42\\
\hline
MS Deformable DETR-R50~\cite{zhu2020deformable}  & 50 & 43.8 & 62.6 & \textbf{47.7} & \textbf{26.4} & 47.1 & 58.0 & 173 & 40\\
MS SMCA-R50~\cite{gao2021fast}  & 50 & 43.7 & 63.6 & 47.2 & 24.2 & 47.0 & 60.4 & 152 & 40\\
MS DETR-GQPos-SiA-R50 & 50 &\textbf{44.0}& \textbf{63.9} & 47.2 & 24.4 & \textbf{47.3} & \textbf{61.2} & 136 & 48 \\
\hline
Faster RCNN-FPN-R50~\cite{ren2015faster}  & 109 & 42.0 & 62.1 & 45.5 & \textbf{26.6} & 45.4 & 53.4 & 180 & 42\\
DETR-R50~\cite{carion2020end}  & 150 & 39.5 & 60.3 & 41.4 & 17.5 & 43.0 & 59.1 & 86 & 41 \\
DETR-R50~\cite{carion2020end}  & 500 & 42.0 & 62.4 & 44.2 & 20.5 & 45.8 & 61.1 & 86 & 41  \\
DETR-DC5-R50~\cite{carion2020end} & 500 & 43.3 & 63.1 & 45.9 & 22.5 & \textbf{47.3} & 61.1 & 187 & 41  \\
UP-DETR-R50~\cite{dai2020up}  & 150 & 40.5 & 60.8 & 42.6 & 19.0 & 44.4 & 60.0 & 86 & 41\\
UP-DETR-R50~\cite{dai2020up}  & 300 & 42.8 &63.0 & 45.3 & 20.8 & 47.1 & 61.7 & 86 & 41\\
SMCA-R50~\cite{gao2021fast}  & 108 & 42.7 & * & * & 22.8 & 46.1 & 60.0 & 86 & 42\\
DETR-GQPos-R50& 108 & \textbf{43.4} &\textbf{64.6} &\textbf{46.1} & 23.1 & \textbf{47.1} & \textbf{61.9} & 86 & 42\\
\hline
\end{tabular}
\end{center}
\caption{DETR's improvements with the help of GQPos and SiA. And we compare it with other DETR related object detectors on COCO 2017 validation set. With our methods, DETR outperforms previous methods. "MS" presents multi-scale}
\label{detr_com}
\end{table*}

\subsection{Object Detection}
\label{object}
\subsubsection{Performance}

\begin{table}[h]
\begin{center}
\begin{tabular}{|l|c|c|c|}
\hline
Methods & Epoch  & mAP\\
\hline\hline
SMCA & 50 & 41.0    \\
SMCA-GQPos & 50 & \textbf{42.5}  \\
\hline
Multi-Scale SMCA & 50 & 43.7 \\
Multi-Scale SMCA-GQPos & 50 & \textbf{44.6}  \\
\hline
\end{tabular}
\end{center}
\caption{Performance improvements on SMCA, the original single and multi-scale SMCA's performance can be further improved by GPos.}
\label{smca:m}
\end{table}

\noindent \textbf{DETR} The results are shown in Table~\ref{detr_com}. With the help of GQPos, DETR's performance can surpass the previous related models whether on the short or long training schedule. Specifically, DETR with GQPos can achieve $42.0$\% mAP at $50$ epochs, exceeding local attention models like ~\cite{gao2021fast} (by $1.6$\% mAP) and ~\cite{zhu2020deformable} (by $1.0$\% mAP) which have a lower complexity for attention calculation, while at $108$ epochs, DETR-GQPos surpasses the performance of its original version at $500$ epochs by $1.4$\% mAP, and exceeds the performance of self-pretrained UP-DETR at $300$ epochs by $0.6$\% mAP. On the other hand, origin DETR's training schedule is reduced to a great extent (DETR-GQPos with $42.0$\% mAP at $50$ epochs versus origin DETR with $42.0$\% mAP at 500 epochs). Above results show that guiding and updating the query position for each cross-attention indeed helps transformer-based detection heads converge faster. \\
Furthermore, with the help of SiA, the performance of DETR-GQPos is further improved by $2$\% mAP. Specifically, the detection performance of the small object is improved by $1.7$\% mAP, indicating the learning of high-resolution attention weight map is accelerated. And when compared with local attention based model, DETR-GQPos-SiA also has an advantage in large object detection, especially over MS Deformable-DETR by $3.2$\% mAP.\\

\begin{table}[t]
\begin{center}
\begin{tabular}{|l|c|c|c|}
\hline
Methods & Epoch  & mAP\\
\hline\hline
YoloS-Ti & 300 & 28.6    \\
YoloS-Ti & 50 & 18.8  \\
Modified-YoloS-Ti & 50 & 24.2 \\
Modified-YoloS-Ti-GQPos & 50 & 25.3  \\
\hline
\end{tabular}
\end{center}
\caption{Performance improvements on YoloS-Ti. "Modified" means we add more $6$ layers and auxiliary losses to origin YoloS-Ti model.}
\label{yolos}
\end{table}

\begin{table*}[h]
\begin{center}
\begin{tabular}{|l|c|c|c|c|c|}
\hline
Methods & Epoch  & Multi-scale inputs & feature maps fusion & attention weight fusion & mAP\\
\hline\hline
DETR-GQPos & 50 & - & - & - & 42.0    \\
DETR-GQPos & 50 & $\surd$ & - & - & 42.1     \\
DETR-GQPos & 50 & $\surd$ & $\surd$ & - & 42.5  \\
DETR-GQPos & 50 & $\surd$ & $\surd$ & $\surd$ & \textbf{44.0}  \\
\hline
\end{tabular}
\end{center}
\caption{Ablations on Similar Attention (SiA) components. }
\label{sia:c}
\end{table*}

\begin{table*}[h]
\begin{center}
\begin{tabular}{|l|c|c|c|c|c|}
\hline
Methods & Epoch  & mAP full & mAP rare & mAP nonrare & mAP inter\\
\hline\hline
HoiTransformer & 30 & 18.01 & 12.03 & 19.79 & 19.63  \\
HoiTransformer-GQPos & 30 & \textbf{18.95} & \textbf{12.13} & \textbf{20.98} &\textbf{20.68}\\
\hline
Multi-scale HoiTransformer & 50 & 20.81 & 13.70 & 22.93 & 22.68 \\
Multi-scale HoiTransformer-SiA & 50 & \textbf{22.10} & \textbf{15.35} &\textbf{24.12} &\textbf{24.06} \\
\hline
\end{tabular}
\end{center}
\caption{Performance improvements on HoiTransformer with GQPos and Similar Attention.}
\label{hoi-gqpos}
\end{table*}

\begin{table}[h]
\begin{center}
\begin{tabular}{|l|c|c|c|}
\hline
Methods & Epoch &  mAP\\
\hline\hline
DETR-GQPos & 50 &  42.0\\
w/o iterative updating & 50  & 40.6 (-1.4)  \\
w/o pos encoding & 50  & 39.3 (\textbf{-2.7}) \\
w/o FC in $Q_{pos}$ & 50  &   41.3(-0.7) \\
\hline
\end{tabular}
\end{center}
\caption{Ablations on GQPos components on DETR model.}
\label{components}
\end{table}

\noindent \textbf{SMCA} Results of GQPos on SMCA are shown in Table~\ref{smca:m}. Combing the GQPos, the performance can be further improved by $1.5$\% mAP under single-scale and $0.9$\% mAP under multi-scale, suggesting that GQPos is not conflicted with local attention based model like SMCA, and has a good generalization ability.\\

\noindent \textbf{YoloS} We first make some modifications to the original YoloS-Ti model as mentioned in experimental settings. Then GQPos is applied to the modified YoloS-Ti model. The results are in show Table~\ref{yolos}. The long training schedule of YoloS is decreased to some extent. Based on the results, we conclude that (i) origin YoloS has no detection heads part except a ViT backbone, which hinders the YoloS from transferring to object detection task. Therefore adding more transformer layers with a larger learning rate enables YoloS to have an approximate detection head. (ii) Inspired by DETR, adding auxiliary losses for the last 6 layers can improve the expression of intermediate object queries. (iii) Above two modifications are critical to YoloS' architecture, bring $5.4$\% mAP gain to the original YoloS-Ti. (iv) GQPos further improve the performance by $1.1$\% mAP, indicating the problem of query position's lack of guiding and updating is widespread.\\ 

\subsubsection{Ablations}
\label{ablation}
We then demonstrate our ablation studies of GQPos and SiA. We use DETR model with Resnet-$50$ backbone to perform object detection on the challenging COCO dataset for ablations.\\
\noindent \textbf{Guided Query Position}\quad The ablations include implementing GQPos not in an iterative way, removing pos encoding part, and fully connected layers of query position and spatial position. Results are shown in Table~\ref{components}. Relevant results demonstrate that: (i) The iterative approach outperforms the parallel one by $1.4$\% mAP since it enables query position to be embedded with the latest location information of object queries. (ii) Embedding object queries' location information to query position is the most important (removing it can cause a $2.7$\% performance drop). In other words, the update of query position needs to be guided. (iii) Adding the linear projection to query position can make the update more dynamic to some extent ($0.7$\% mAP improvements).\\
\noindent \textbf{Similar Attention}\quad Results about SiA's ablations are shown in Table~\ref{sia:c}. We observe the effects of adding multi-scale features or fusing multi-scale features are limited ($0.1$\% mAP and $0.5$\% mAP respectively), demonstrating the high complexity problem of high-resolution feature map should be addressed. And when combined with the attention weights fusion part in SiA, the performance obtains a $2.0$\% mAP gain, showing that using the prior of low-resolution feature map is critical to learn the high-resolution feature map.

\subsection{Human Object Interaction}
\label{hoi:t}
The results of applying GQPos and SiA respectively to single and multi-scale HoiTransformer are shown in Table~\ref{hoi-gqpos}. Both of the methods can improve HoiTransformer's performance. By adding GQPos, HoiTransformer obtains a $0.94$\% mAP gain for full objects at $30$ epochs and by adding SiA, and multi-scale HoiTransformer obtains a $1.29$\% mAP gain for full objects at $50$ epochs. In particular, SiA has a significant improvement in the detection of rare objects ($13.70$\% mAP versus $15.35$\% mAP), suggesting that in human-object-interaction, the rare objects detection can be enhanced with well-learned multi-scale information. Overall, GQPos and SiA can generalize well to other vision tasks like human-object-interaction.

\begin{figure*}[h]
\begin{center}
\vbox{
    \begin{subfigure}{\textwidth}
        \centering
        % \hbox{%
        \begin{subfigure}{0.16\textwidth}
            \includegraphics[width=\textwidth]{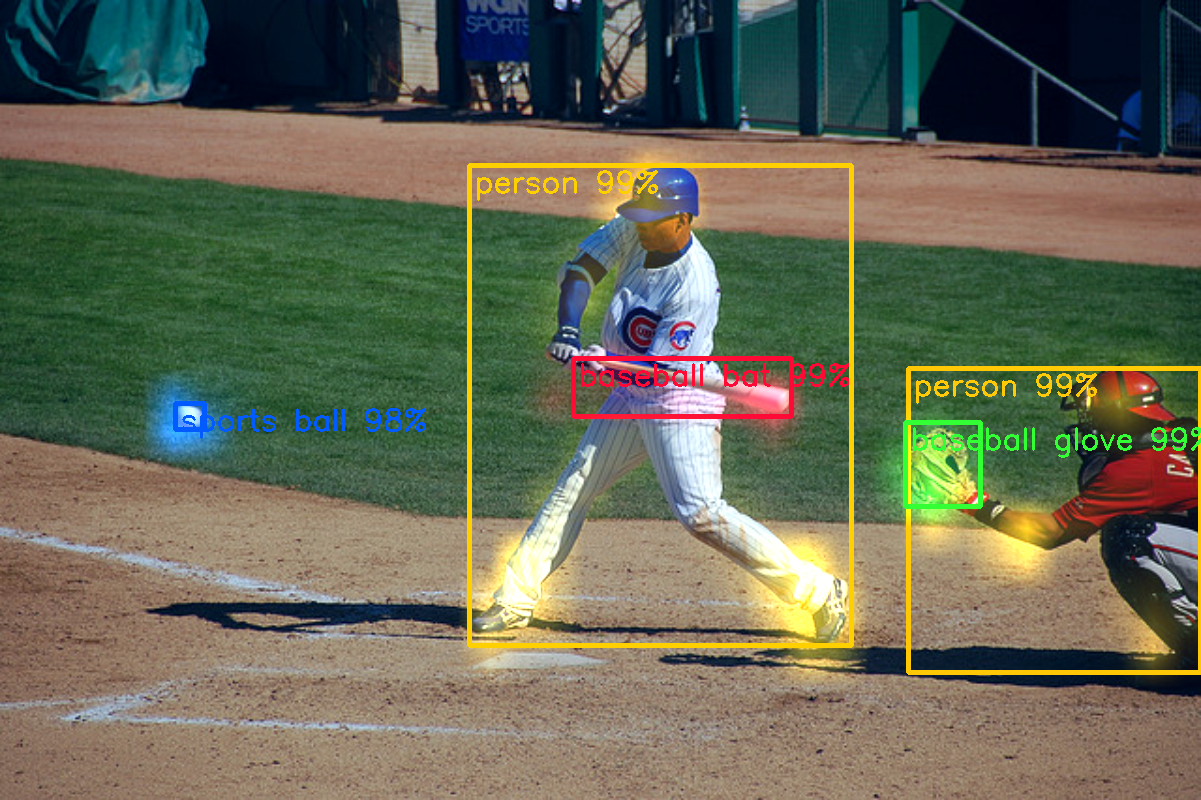}
        \end{subfigure}
        \begin{subfigure}{0.16\textwidth}
            \includegraphics[width=\textwidth]{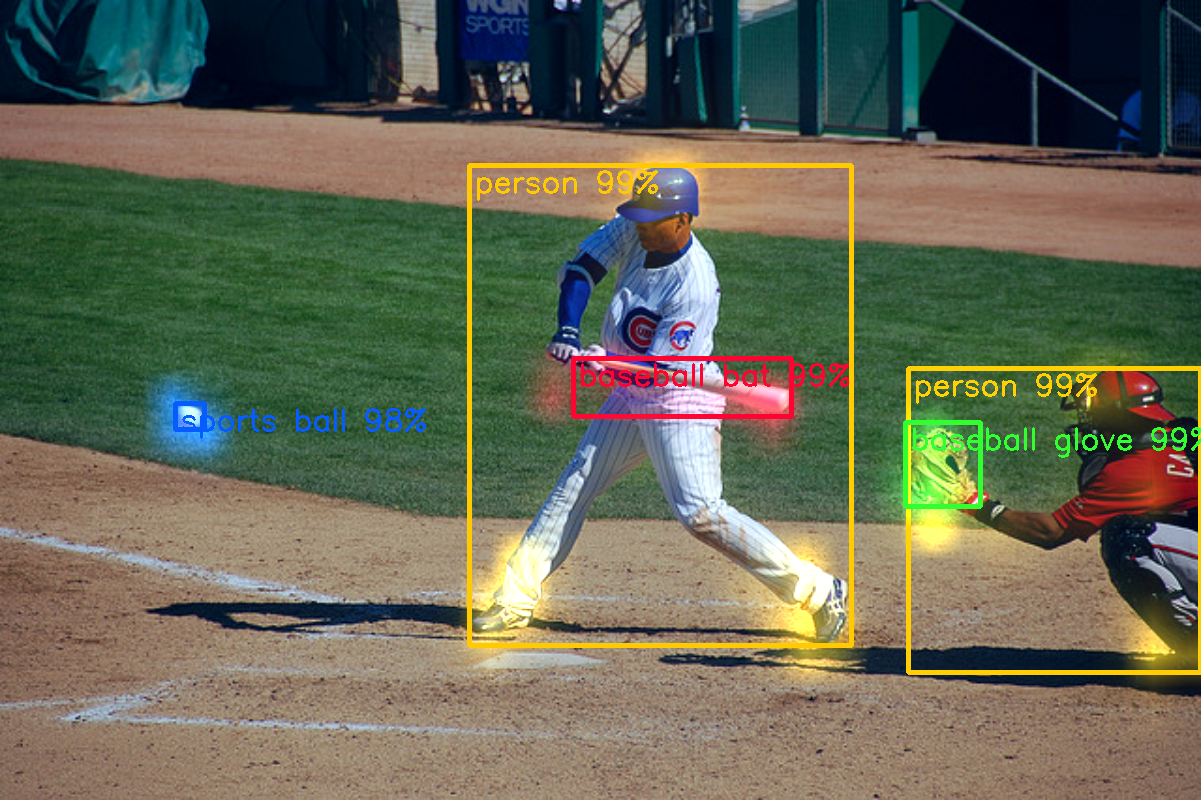}
        \end{subfigure}
        \begin{subfigure}{0.16\textwidth}
            \includegraphics[width=\textwidth]{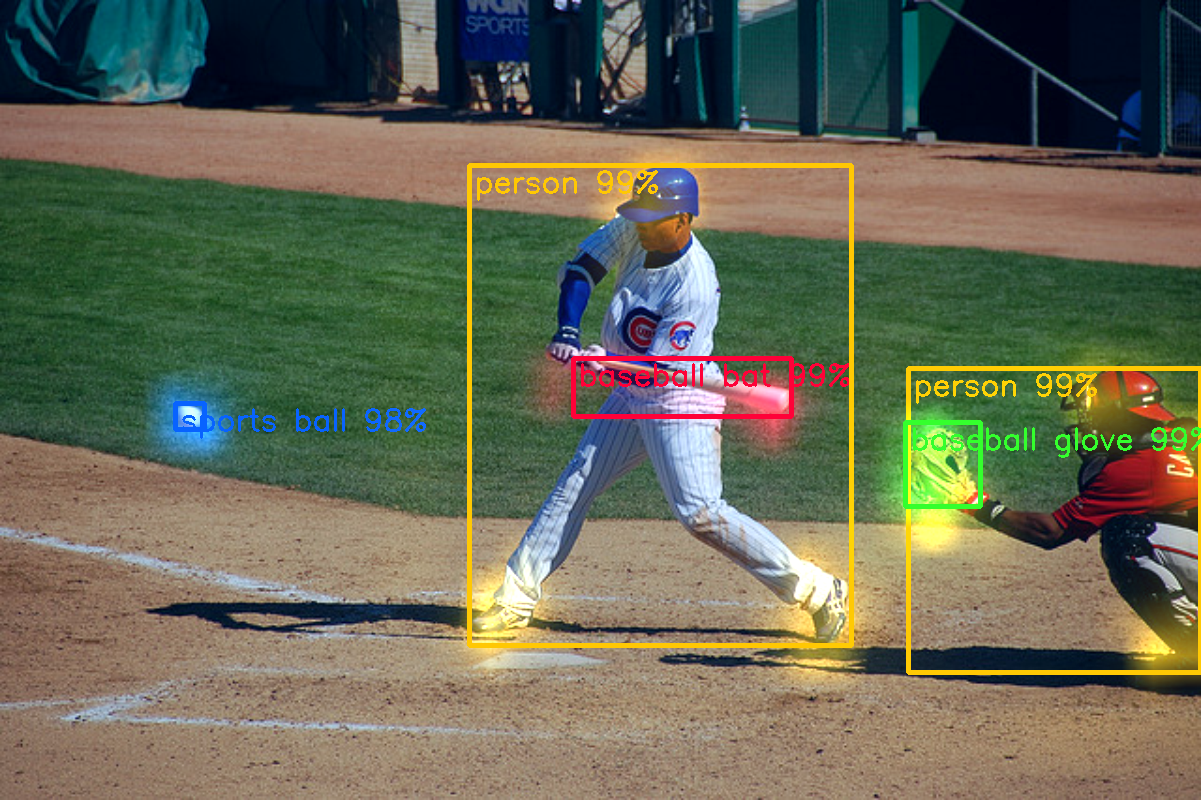}
        \end{subfigure}
        \begin{subfigure}{0.16\textwidth}
            \includegraphics[width=\textwidth]{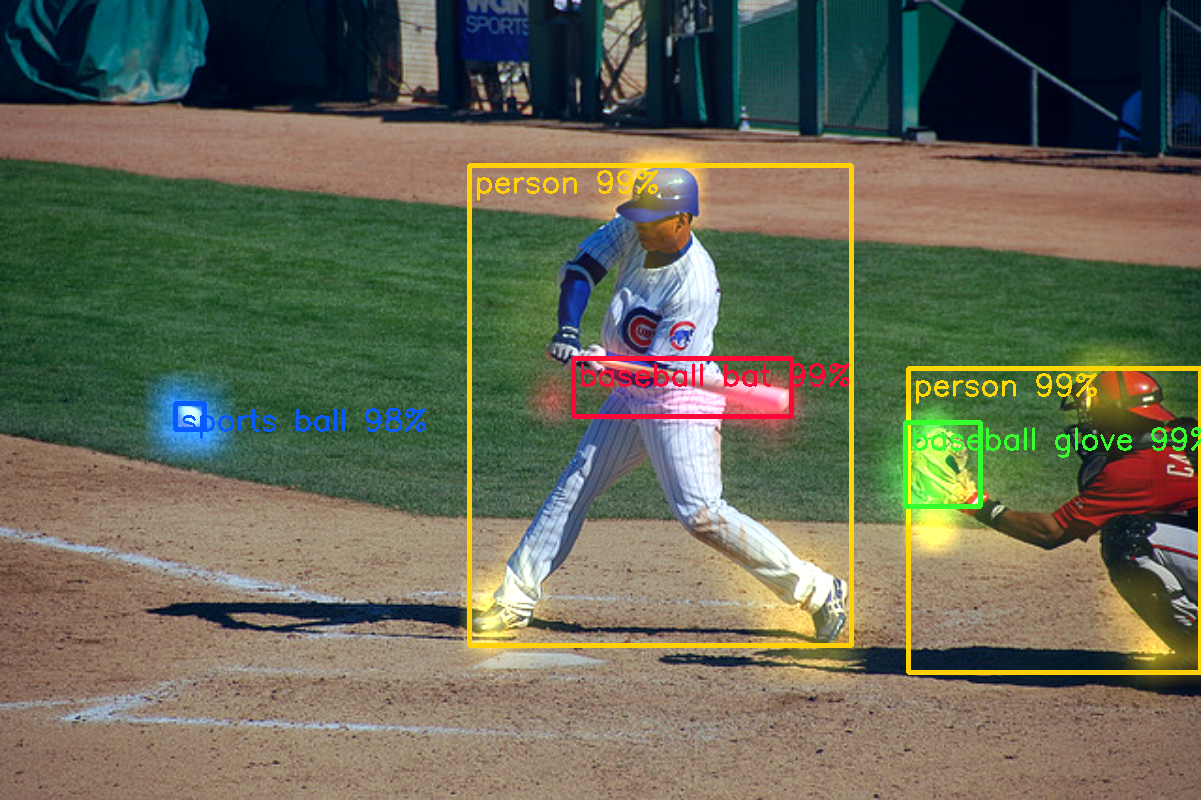}
        \end{subfigure}
        \begin{subfigure}{0.16\textwidth}
            \includegraphics[width=\textwidth]{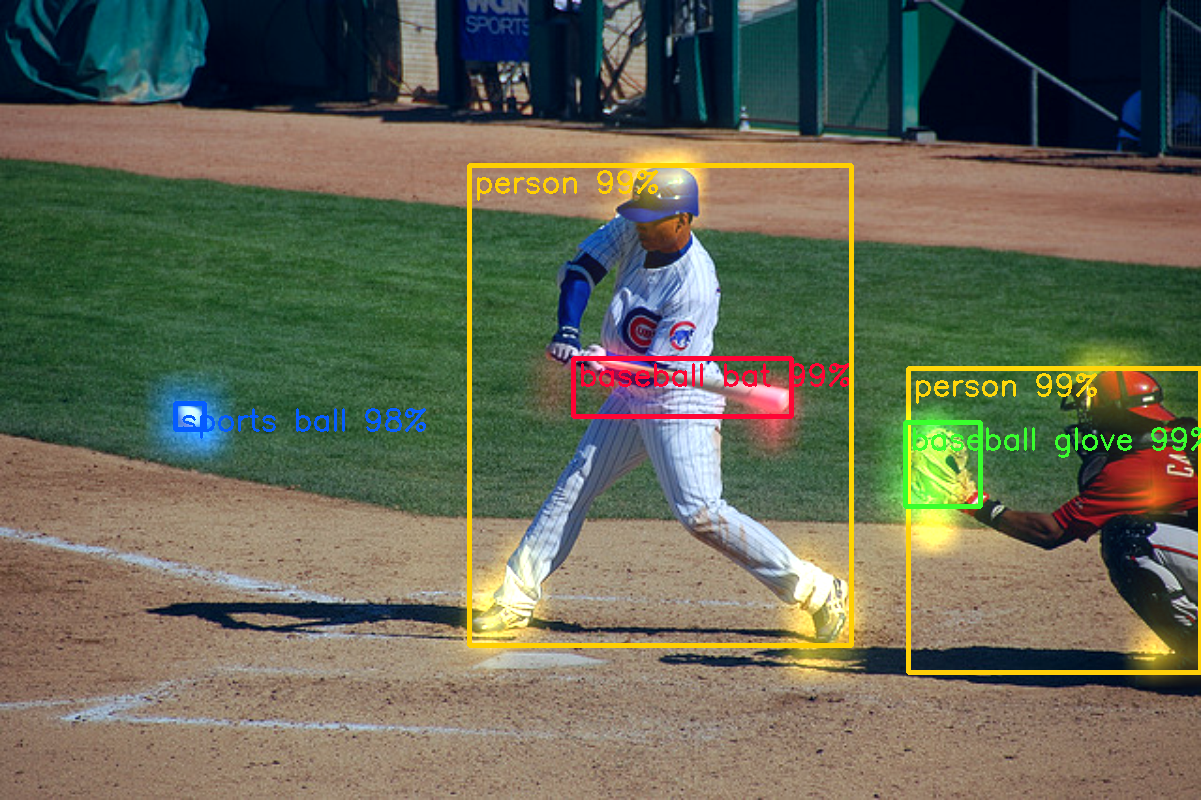}
        \end{subfigure}
        \begin{subfigure}{0.16\textwidth}
            \includegraphics[width=\textwidth]{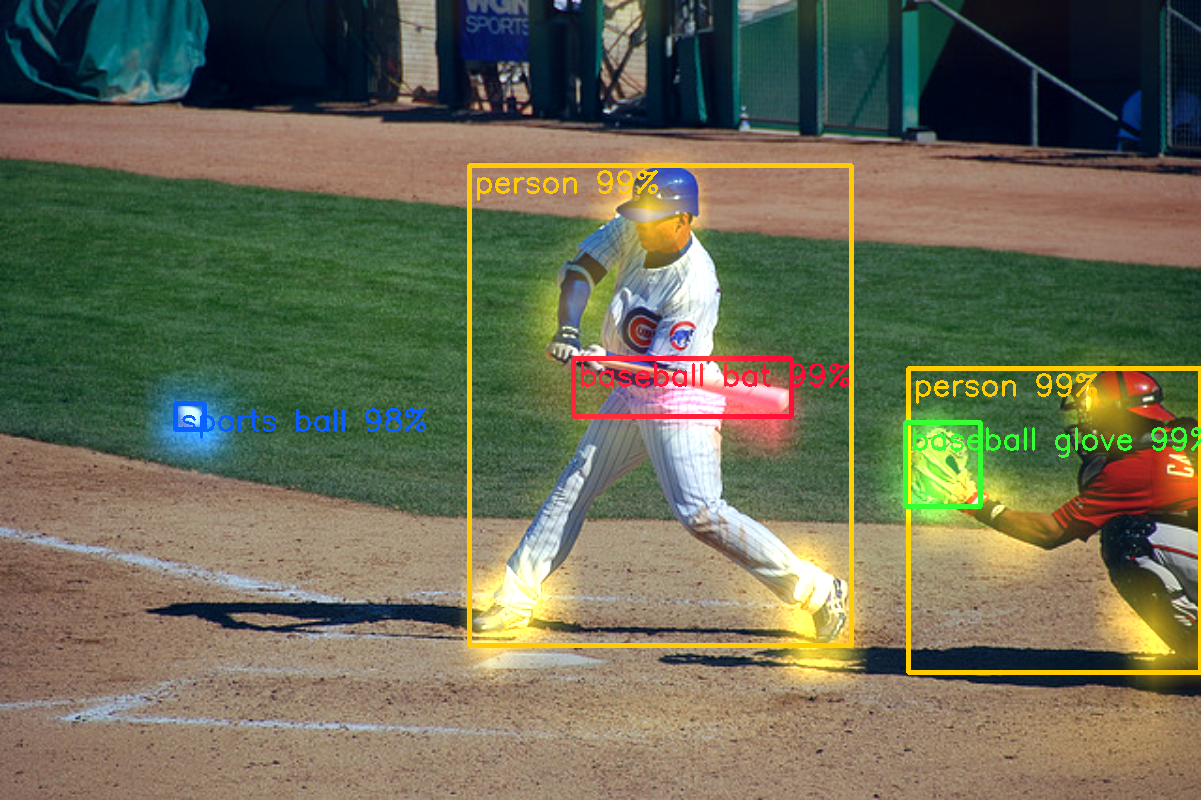}
        \end{subfigure}
        % }
        \caption{DETR}
        \label{fig:vis_detr}
    \end{subfigure}
    \begin{subfigure}{\textwidth}
        \centering
        \begin{subfigure}{0.16\textwidth}
            \includegraphics[width=\textwidth]{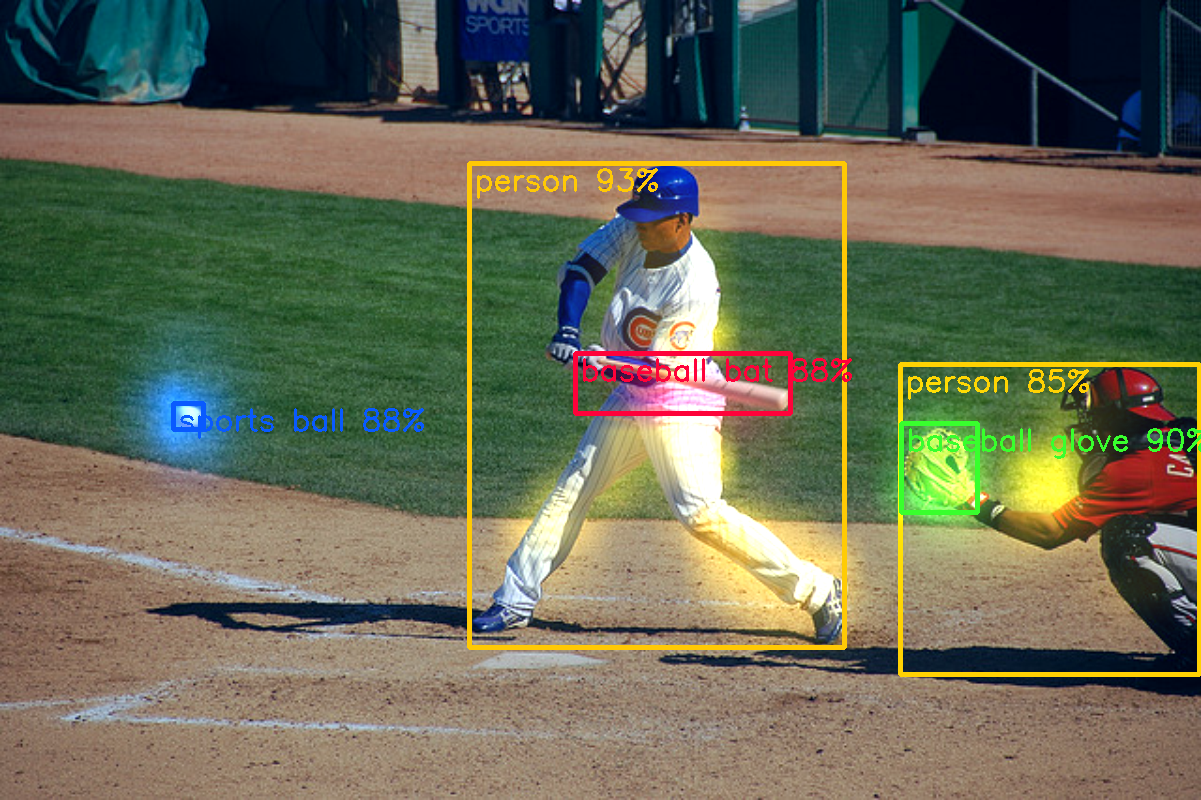}
        \end{subfigure}
        \begin{subfigure}{0.16\textwidth}
            \includegraphics[width=\textwidth]{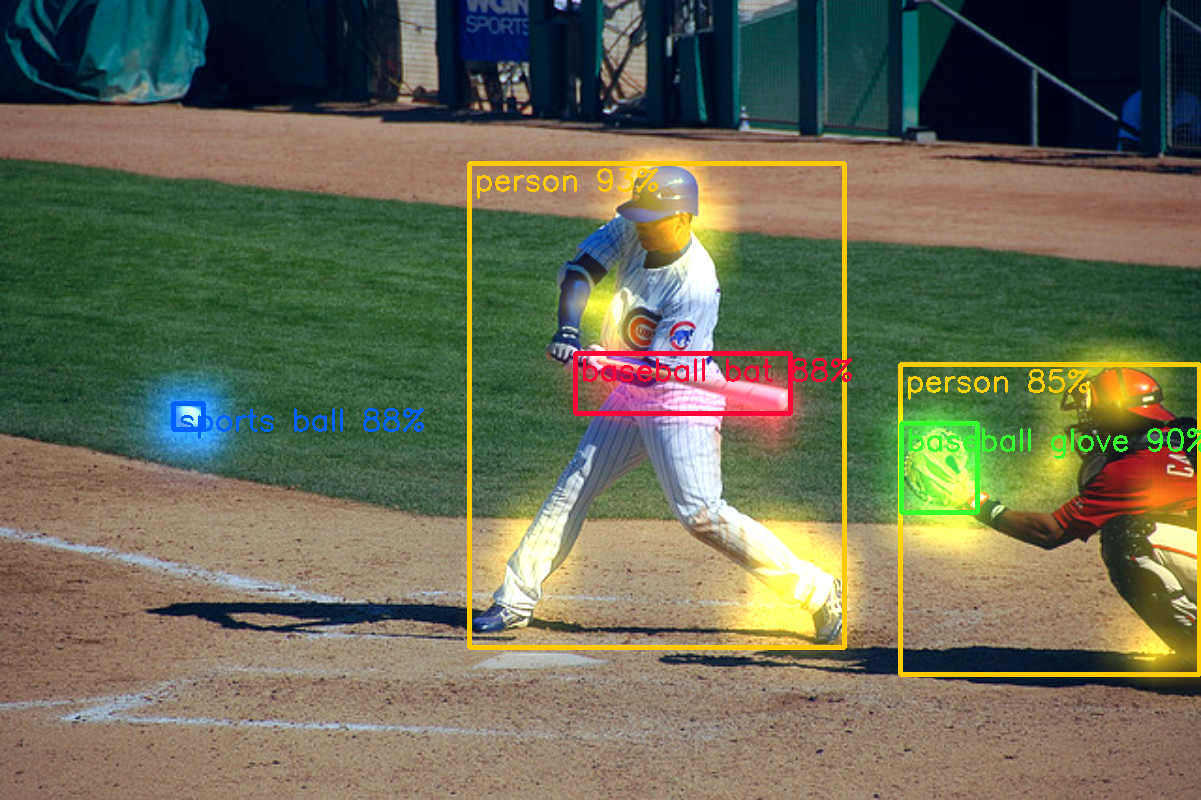}
        \end{subfigure}
        \begin{subfigure}{0.16\textwidth}
            \includegraphics[width=\textwidth]{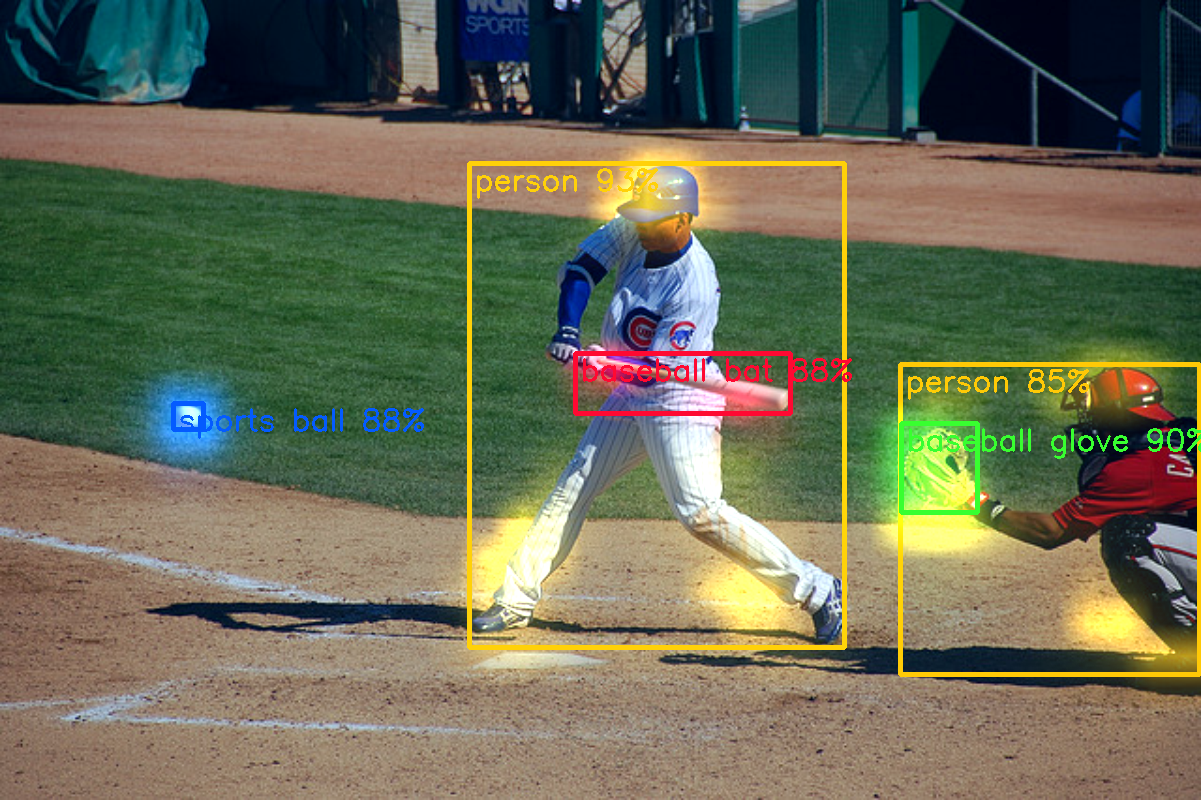}
        \end{subfigure}
        \begin{subfigure}{0.16\textwidth}
            \includegraphics[width=\textwidth]{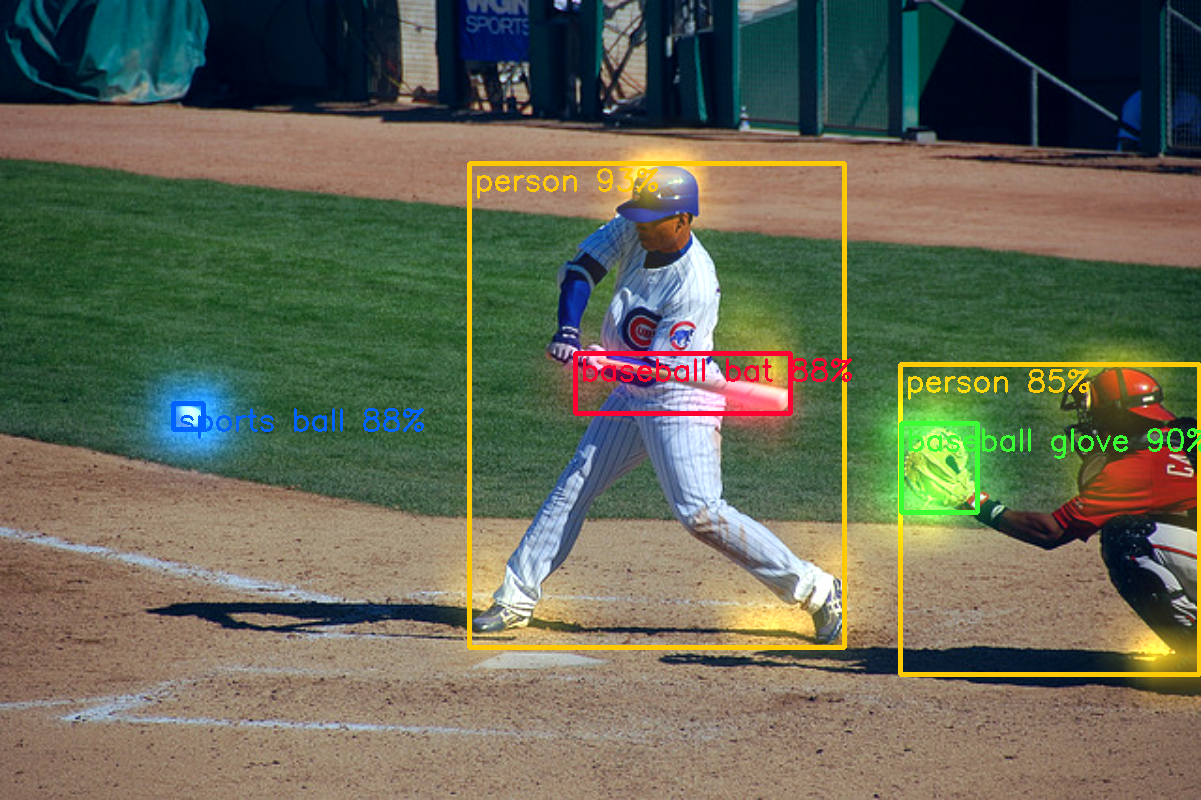}
        \end{subfigure}
        \begin{subfigure}{0.16\textwidth}
            \includegraphics[width=\textwidth]{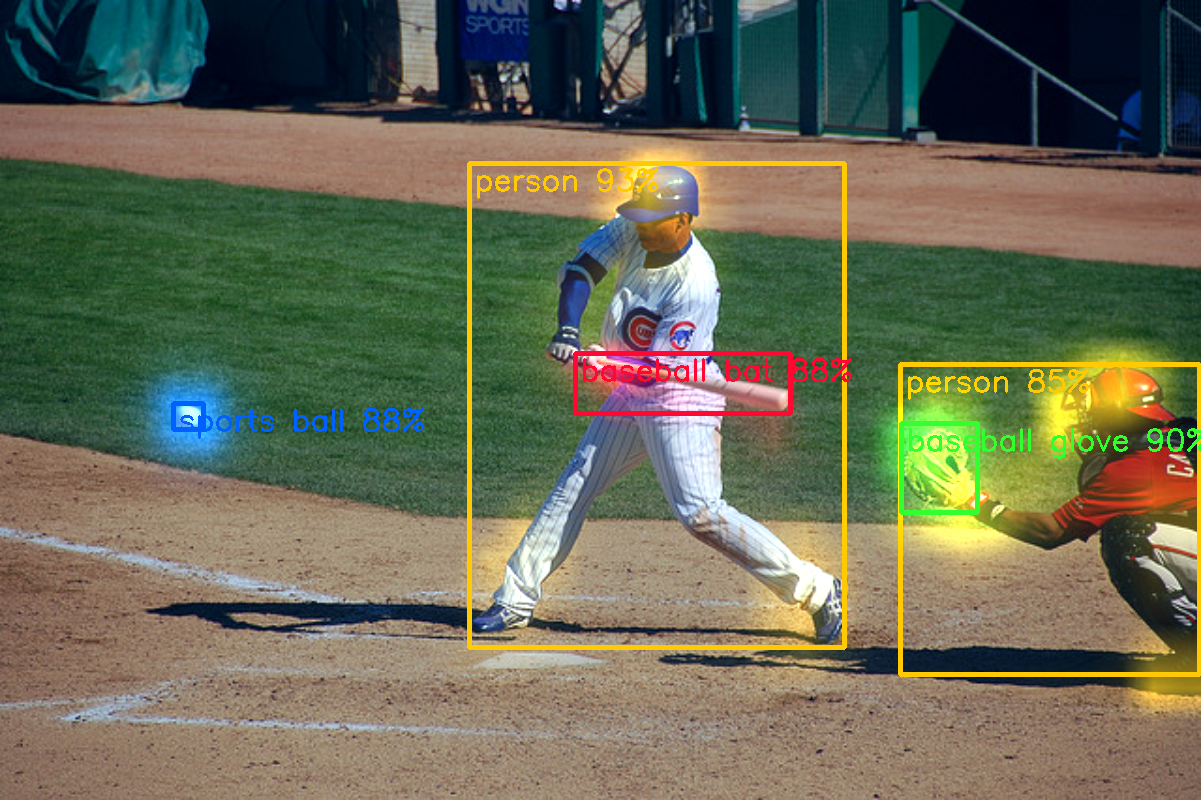}
        \end{subfigure}
        \begin{subfigure}{0.16\textwidth}
            \includegraphics[width=\textwidth]{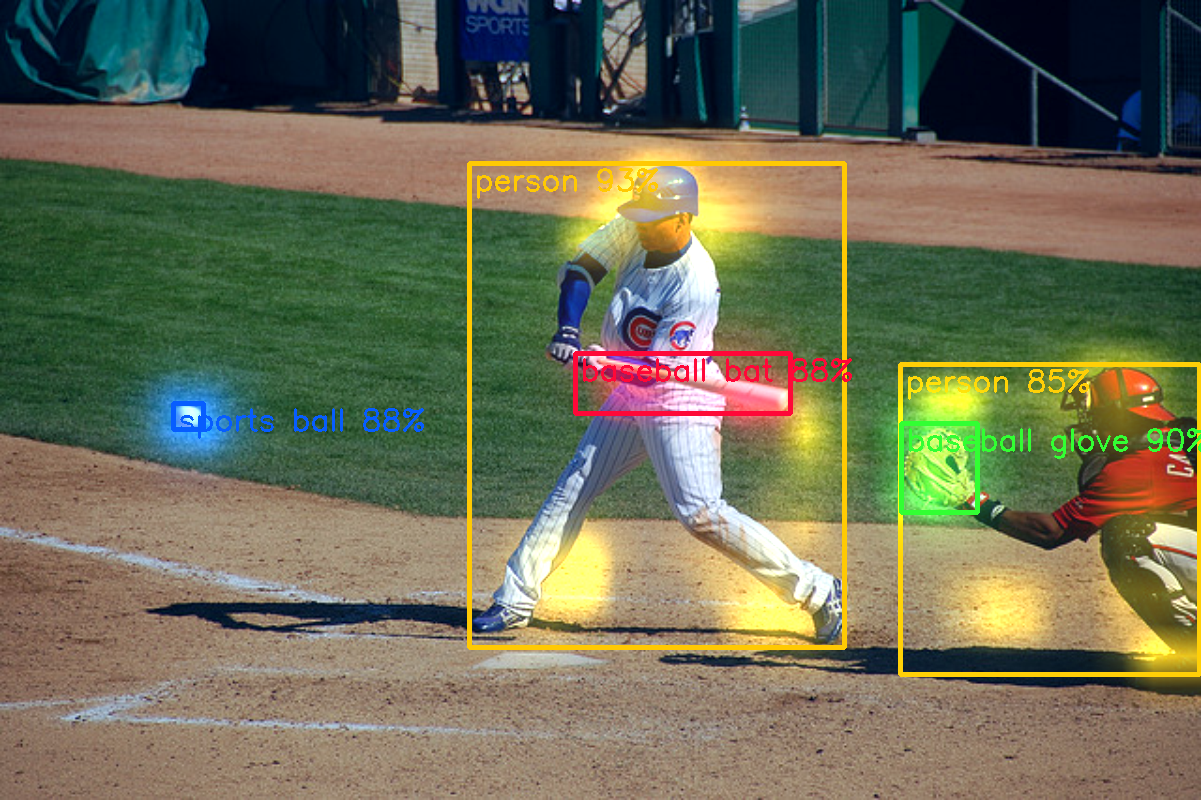}
        \end{subfigure}
        \caption{DETR w/ GQPos}
        \label{fig:vis_iter}
    \end{subfigure}
}
\end{center}
   \caption{Visualization of decoder attention maps of DETR and DETR with GQPos. The $6$ columns represent $6$ decoder layers, and the attention region with high probability is highlighted. DETR tries to find object boundaries from the first decoder layer and attend to similar locations over different layers. With GQPos, our model locates objects more reasonably by first focusing on the object center, and then gradually moving the attention to the boundary region.}
\label{fig:vis}
\end{figure*}

\subsection{Visualization}
\label{visualization}
To explore the inherent mechanism of GQPos, we visualize the decoder feature maps of different layers. As Figure~\ref{fig:vis_detr} shows, the decoder of origin DETR attends to object boundaries for all layers. so that the attention maps keep high consistency among different layers, which means different decoder layers bring little improvement for attention representation.

Interestingly, GQPos locates objects in a more humanoid way. That is, first find the object center, and then locate the boundaries. Figure~\ref{fig:vis_iter} shows the decoder feature maps of GQPos. Unlike origin DETR, the decoder with GQPos attends to the central part of the object at the beginning, such as the human body in Figure~\ref{fig:vis_iter}. Compared to locating object boundary, locating object center is easier and thus the initial query position can be better optimized. After locating objects by central part, the focus of attention maps gradually moves to object boundaries in subsequent layers, which shows the refinement ability of GQPos. 

\section{Conclusion}
We propose that when transformer-based detection heads performing cross-attention, the interaction between query position and feature maps needs to be updated as the object queries are renewed. And a simple and effective method GQPos is introduced. GQPos shows good generalizations and is compatible with DETR, SMCA, YoloS, and HoiTransformer. Furthermore, for Transformer with multi-scale feature maps, we observe that feature maps fusion is not enough and propose SiA method to additionally fuse the multi-scale attention weight maps which accelerate the learning of high-resolution attention weight maps. SiA shows improvements on multi-scale transformer-based detection heads like DETR and HoiTransformer.
{\small
\bibliographystyle{ieee_fullname}
\bibliography{egbib}

\begin{thebibliography}{10}\itemsep=-1pt

\bibitem{carion2020end}
Nicolas Carion, Francisco Massa, Gabriel Synnaeve, Nicolas Usunier, Alexander
  Kirillov, and Sergey Zagoruyko.
\newblock End-to-end object detection with transformers.
\newblock In {\em European Conference on Computer Vision}, pages 213--229.
  Springer, 2020.

\bibitem{chao2015hico}
Yu-Wei Chao, Zhan Wang, Yugeng He, Jiaxuan Wang, and Jia Deng.
\newblock Hico: A benchmark for recognizing human-object interactions in
  images.
\newblock In {\em Proceedings of the IEEE International Conference on Computer
  Vision}, pages 1017--1025, 2015.

\bibitem{chen2020pre}
Hanting Chen, Yunhe Wang, Tianyu Guo, Chang Xu, Yiping Deng, Zhenhua Liu, Siwei
  Ma, Chunjing Xu, Chao Xu, and Wen Gao.
\newblock Pre-trained image processing transformer.
\newblock {\em arXiv preprint arXiv:2012.00364}, 2020.

\bibitem{dai2020up}
Zhigang Dai, Bolun Cai, Yugeng Lin, and Junying Chen.
\newblock Up-detr: Unsupervised pre-training for object detection with
  transformers.
\newblock {\em arXiv preprint arXiv:2011.09094}, 2020.

\bibitem{dosovitskiy2020image}
Alexey Dosovitskiy, Lucas Beyer, Alexander Kolesnikov, Dirk Weissenborn,
  Xiaohua Zhai, Thomas Unterthiner, Mostafa Dehghani, Matthias Minderer, Georg
  Heigold, Sylvain Gelly, et~al.
\newblock An image is worth 16x16 words: Transformers for image recognition at
  scale.
\newblock {\em arXiv preprint arXiv:2010.11929}, 2020.

\bibitem{fang2021you}
Yuxin Fang, Bencheng Liao, Xinggang Wang, Jiemin Fang, Jiyang Qi, Rui Wu,
  Jianwei Niu, and Wenyu Liu.
\newblock You only look at one sequence: Rethinking transformer in vision
  through object detection.
\newblock {\em arXiv preprint arXiv:2106.00666}, 2021.

\bibitem{gao2021fast}
Peng Gao, Minghang Zheng, Xiaogang Wang, Jifeng Dai, and Hongsheng Li.
\newblock Fast convergence of detr with spatially modulated co-attention.
\newblock {\em arXiv preprint arXiv:2101.07448}, 2021.

\bibitem{girdhar2019video}
Rohit Girdhar, Joao Carreira, Carl Doersch, and Andrew Zisserman.
\newblock Video action transformer network.
\newblock In {\em Proceedings of the IEEE/CVF Conference on Computer Vision and
  Pattern Recognition}, pages 244--253, 2019.

\bibitem{girshick2015fast}
Ross Girshick.
\newblock Fast r-cnn.
\newblock In {\em Proceedings of the IEEE international conference on computer
  vision}, pages 1440--1448, 2015.

\bibitem{girshick2015region}
Ross Girshick, Jeff Donahue, Trevor Darrell, and Jitendra Malik.
\newblock Region-based convolutional networks for accurate object detection and
  segmentation.
\newblock {\em IEEE transactions on pattern analysis and machine intelligence},
  38(1):142--158, 2015.

\bibitem{he2016deep}
Kaiming He, Xiangyu Zhang, Shaoqing Ren, and Jian Sun.
\newblock Deep residual learning for image recognition.
\newblock In {\em Proceedings of the IEEE conference on computer vision and
  pattern recognition}, pages 770--778, 2016.

\bibitem{krizhevsky2012imagenet}
Alex Krizhevsky, Ilya Sutskever, and Geoffrey~E Hinton.
\newblock Imagenet classification with deep convolutional neural networks.
\newblock {\em Advances in neural information processing systems},
  25:1097--1105, 2012.

\bibitem{lin2017feature}
Tsung-Yi Lin, Piotr Doll{\'a}r, Ross Girshick, Kaiming He, Bharath Hariharan,
  and Serge Belongie.
\newblock Feature pyramid networks for object detection.
\newblock In {\em Proceedings of the IEEE conference on computer vision and
  pattern recognition}, pages 2117--2125, 2017.

\bibitem{lin2017focal}
Tsung-Yi Lin, Priya Goyal, Ross Girshick, Kaiming He, and Piotr Doll{\'a}r.
\newblock Focal loss for dense object detection.
\newblock In {\em Proceedings of the IEEE international conference on computer
  vision}, pages 2980--2988, 2017.

\bibitem{lin2014microsoft}
Tsung-Yi Lin, Michael Maire, Serge Belongie, James Hays, Pietro Perona, Deva
  Ramanan, Piotr Doll{\'a}r, and C~Lawrence Zitnick.
\newblock Microsoft coco: Common objects in context.
\newblock In {\em European conference on computer vision}, pages 740--755.
  Springer, 2014.

\bibitem{liu2016ssd}
Wei Liu, Dragomir Anguelov, Dumitru Erhan, Christian Szegedy, Scott Reed,
  Cheng-Yang Fu, and Alexander~C Berg.
\newblock Ssd: Single shot multibox detector.
\newblock In {\em European conference on computer vision}, pages 21--37.
  Springer, 2016.

\bibitem{loshchilov2018fixing}
Ilya Loshchilov and Frank Hutter.
\newblock Fixing weight decay regularization in adam.
\newblock 2018.

\bibitem{mikolov2010recurrent}
Tom{\'a}{\v{s}} Mikolov, Martin Karafi{\'a}t, Luk{\'a}{\v{s}} Burget, Jan
  {\v{C}}ernock{\`y}, and Sanjeev Khudanpur.
\newblock Recurrent neural network based language model.
\newblock In {\em Eleventh annual conference of the international speech
  communication association}, 2010.

\bibitem{redmon2016you}
Joseph Redmon, Santosh Divvala, Ross Girshick, and Ali Farhadi.
\newblock You only look once: Unified, real-time object detection.
\newblock In {\em Proceedings of the IEEE conference on computer vision and
  pattern recognition}, pages 779--788, 2016.

\bibitem{ren2015faster}
Shaoqing Ren, Kaiming He, Ross Girshick, and Jian Sun.
\newblock Faster r-cnn: Towards real-time object detection with region proposal
  networks.
\newblock {\em arXiv preprint arXiv:1506.01497}, 2015.

\bibitem{stewart2016end}
Russell Stewart, Mykhaylo Andriluka, and Andrew~Y Ng.
\newblock End-to-end people detection in crowded scenes.
\newblock In {\em Proceedings of the IEEE conference on computer vision and
  pattern recognition}, pages 2325--2333, 2016.

\bibitem{su2019vl}
Weijie Su, Xizhou Zhu, Yue Cao, Bin Li, Lewei Lu, Furu Wei, and Jifeng Dai.
\newblock Vl-bert: Pre-training of generic visual-linguistic representations.
\newblock {\em arXiv preprint arXiv:1908.08530}, 2019.

\bibitem{sun2019videobert}
Chen Sun, Austin Myers, Carl Vondrick, Kevin Murphy, and Cordelia Schmid.
\newblock Videobert: A joint model for video and language representation
  learning.
\newblock In {\em Proceedings of the IEEE/CVF International Conference on
  Computer Vision}, pages 7464--7473, 2019.

\bibitem{sun2020rethinking}
Zhiqing Sun, Shengcao Cao, Yiming Yang, and Kris Kitani.
\newblock Rethinking transformer-based set prediction for object detection.
\newblock {\em arXiv preprint arXiv:2011.10881}, 2020.

\bibitem{tan2019lxmert}
Hao Tan and Mohit Bansal.
\newblock Lxmert: Learning cross-modality encoder representations from
  transformers.
\newblock {\em arXiv preprint arXiv:1908.07490}, 2019.

\bibitem{tian2019fcos}
Zhi Tian, Chunhua Shen, Hao Chen, and Tong He.
\newblock Fcos: Fully convolutional one-stage object detection.
\newblock In {\em Proceedings of the IEEE/CVF International Conference on
  Computer Vision}, pages 9627--9636, 2019.

\bibitem{touvron2020training}
Hugo Touvron, Matthieu Cord, Matthijs Douze, Francisco Massa, Alexandre
  Sablayrolles, and Herv{\'e} J{\'e}gou.
\newblock Training data-efficient image transformers \& distillation through
  attention.
\newblock {\em arXiv preprint arXiv:2012.12877}, 2020.

\bibitem{vaswani2017attention}
Ashish Vaswani, Noam Shazeer, Niki Parmar, Jakob Uszkoreit, Llion Jones,
  Aidan~N Gomez, Lukasz Kaiser, and Illia Polosukhin.
\newblock Attention is all you need.
\newblock {\em arXiv preprint arXiv:1706.03762}, 2017.

\bibitem{wang2020end}
Jianfeng Wang, Lin Song, Zeming Li, Hongbin Sun, Jian Sun, and Nanning Zheng.
\newblock End-to-end object detection with fully convolutional network.
\newblock {\em arXiv preprint arXiv:2012.03544}, 2020.

\bibitem{yang2020learning}
Fuzhi Yang, Huan Yang, Jianlong Fu, Hongtao Lu, and Baining Guo.
\newblock Learning texture transformer network for image super-resolution.
\newblock In {\em Proceedings of the IEEE/CVF Conference on Computer Vision and
  Pattern Recognition}, pages 5791--5800, 2020.

\bibitem{ye2019cross}
Linwei Ye, Mrigank Rochan, Zhi Liu, and Yang Wang.
\newblock Cross-modal self-attention network for referring image segmentation.
\newblock In {\em Proceedings of the IEEE/CVF Conference on Computer Vision and
  Pattern Recognition}, pages 10502--10511, 2019.

\bibitem{zheng2020end}
Minghang Zheng, Peng Gao, Xiaogang Wang, Hongsheng Li, and Hao Dong.
\newblock End-to-end object detection with adaptive clustering transformer.
\newblock {\em arXiv preprint arXiv:2011.09315}, 2020.

\bibitem{zhu2020deformable}
Xizhou Zhu, Weijie Su, Lewei Lu, Bin Li, Xiaogang Wang, and Jifeng Dai.
\newblock Deformable detr: Deformable transformers for end-to-end object
  detection.
\newblock {\em arXiv preprint arXiv:2010.04159}, 2020.

\bibitem{zou2021end}
Cheng Zou, Bohan Wang, Yue Hu, Junqi Liu, Qian Wu, Yu Zhao, Boxun Li, Chenguang
  Zhang, Chi Zhang, Yichen Wei, et~al.
\newblock End-to-end human object interaction detection with hoi transformer.
\newblock In {\em Proceedings of the IEEE/CVF Conference on Computer Vision and
  Pattern Recognition}, pages 11825--11834, 2021.

\end{thebibliography}
}

\end{document}